\documentclass{article} 
\usepackage{arxiv}

\usepackage[utf8]{inputenc} 
\usepackage[T1]{fontenc}    
\usepackage{hyperref}       
\usepackage{url}            
\usepackage{booktabs}       
\usepackage{amsfonts}       
\usepackage{nicefrac}       
\usepackage{microtype}      
\usepackage{lipsum}		

\usepackage{graphicx}
\usepackage{subfig}
\usepackage{amsmath}
\usepackage{graphicx}
\usepackage{algpseudocode}
\usepackage{algorithm}  
\usepackage{algorithmicx} 
\usepackage{algpseudocode} 
\usepackage{amssymb}
\usepackage{amsmath}
\usepackage{bbding}
\usepackage{multirow}
\usepackage{setspace}

\title{Depth Edge Guided CNNs for Sparse Depth Upsampling}

\author{
  Yi Guo \\
  College of Computer Science\\
  Chongqing University\\
  Chongqing 400044, China \\
  \texttt{masterguo@cqu.edu.cn} \\
   \And
  Ji Liu \\
  College of Computer Science\\
  Chongqing University\\
  Chongqing 400044, China \\
  \texttt{liujiboy@cqu.edu.cn} \\
}

\begin{document}
\maketitle
\abstract{Guided sparse depth upsampling aims to upsample an irregularly sampled sparse depth map when an aligned high-resolution color image is given as guidance. Many neural networks have been designed for this task. However, they often ignore the structural difference between the depth and the color image, resulting in obvious artifacts such as texture copy and depth blur at the upsampling depth. Inspired by the normalized convolution operation, we propose a guided convolutional layer to recover dense depth from sparse and irregular depth image with an depth edge image as guidance. Our novel guided network can prevent the depth value from crossing the depth edge to facilitate upsampling. We further design a convolution network based on proposed convolutional layer to combine the advantages of different algorithms and achieve better performance. We conduct comprehensive experiments to verify our method on real-world  indoor and synthetic outdoor datasets. Our method produces strong results. It outperforms state-of-the-art methods on the Virtual KITTI dataset and the Middlebury dataset. It also presents strong generalization capability under different 3D point densities, various lighting and weather conditions.}


\section{Introduction}
The purpose of sparse depth upsampling is to reconstruct dense depth maps from sparsely measured data that is not regularly sampled. Since the combination of 3D laser scanners and cameras became popular in autonomous driving, this task has received considerable attention. Due to the limitation of hardware development, the most advanced range sensors today still achieve much lower resolution data than visual images. Even for the Velodyne HDL-64e \cite{Velodyne2018}, when a sparse 3D point cloud is projected onto an aligned 2D image, it only obtains about 5\% effective depth values in the projected image. Such a high sparsity level makes it challenging to perform subsequent tasks such as RGB-D based object detection\cite{yang2019std,liang2019multi,du2018general,qi2018frustum} and road scene understanding\cite{caltagirone2019lidar,chen2019progressive}.

While some methods\cite{riegler2016atgv,uhrig2017sparsity}operate directly on depth input for upsampling, others\cite{tomasi1998bilateral,petschnigg2004digital,diebel2006application,park2011high,riegler2016deep,hui2016depth,li2016deep} require guidance, for example, from high-resolution images, and we'll focus here. The basic assumption used to guide upsampling is that the target domain shares commonality with high-resolution guided images, for example, image edges are aligned with depth discontinuities. A popular choice for guided upsampling is guided bilateral filtering\cite{chan2008noise,dolson2010upsampling,kopf2007joint,liu2013joint,yang2007spatial}. More advanced methods are based on global energy minimization\cite{barron2016fast,diebel2006application,ferstl2013image,park2011high,riegler2016deep}, compressed sensing\cite{hawe2011dense}or combined semantics to improve performance\cite{schneider2016semantically}. Some methods also use the end-to-end model to perform guided depth upsampling of regular data\cite{hui2016depth,song2016deep}.

Guided sparse depth upsampling methods are usually based on the assumption that depth edges consistently appear at the edges of the images. However, despite there are close connections between depth map and image, the discontinuities—especially the textures—of image are not always consistent with those of depth map. Using the color image as guidance, they ignore the structural difference between the depth and the guidance color image, resulting in obvious artifacts such as texture copy and depth blur at the upsampling depth. 

We propose a novel sparse depth upsampling convolution layer(EGCL) using depth edge maps to better perform the  sparse depth upsampling task. Different from previous methods, we transform the problem of sparse depth upsampling from color image guidance to depth edge guidance, because depth edges have special importance in non-textured depth images. Under the guidance of predicting the depth edge map, an improved confidence propagation CNNs\cite{eldesokey2019confidence} method is used to reconstruct the dense depth image. Depth edge guidance not only helps to avoid artifacts caused by direct image prediction, but also reduces aliasing artifacts and preserves sharp edges. Experiments on various datasets demonstrate that our approach outperforms the state-of-the-arts. 

Based upon EGCL, we design a new architecture named edge guided convolutional neural network (EGCNN) to perform guided sparse depth upsampling. EGCNN consists of three major components: edge information extraction component, depth upsampling subnetwork, and depth fusion subnetwork. Edge information extraction component takes color images as input and outputs their edge-dist fields. Depth upsampling subnetwork applies EGCLs to deal with a sparse depth map with the guidance of edge-dist field. Depth fusion subnetwork is designed to fuse the outputs of depth upsampling subnetwork together to predict the dense upsampling result. Experiments show that this architecture can further improve performance.

\section{Related Work}
Since the emergence of active sensors with depth capabilities, depth completion has become a fundamental task in computer vision. In this section, we review previous literatures on this topic. Depending on whether there is an RGB image to guide the depth completion, previous methods can be roughly divided into two categories: non-guided methods and guided methods. We briefly review these techniques.
\subsection{Non-guided Depth Upsampling}
Methods for non-guided depth upsampling are closely related to those for single image super-resolution. Early methods were usually based on interpolation\cite{hou1978cubic}, sparse representation\cite{yang2010image}, and other traditional techniques. Recently, methods utilizing deep learning have achieved great success in both deep\cite{riegler2016atgv,uhrig2017sparsity}and color\cite{dong2015image,kim2016accurate}image super-resolution. Dahl et al.\cite{dahl2017pixel}presented a pixel recursive super resolution model that synthesizes realistic details into images while enhancing their resolution. Ma et al.\cite{ma2019self} utilized an encoder-decoder architecture with self-supervised framework to predict the dense output. All of the above methods can handle regular low-resolution images but are not suitable for sparse data with irregular sampling.

To pay special attention to irregular sparse data, Chodosh et al.\cite{chodosh2018deep}used compressive sensing to handle sparsity, while using binary masks to filter out missing values. Uhrig et al.\cite{uhrig2017sparsity}proposed a sparse invariant convolution layer, which uses a binary effective mask to normalize sparse inputs. This layer is used to train a sparse depth mapping network with binary effective mask as input and dense depth mapping as output. Similarly, Hua and Gong\cite{hua2018normalized}proposed a similar layer that uses a trained convolution filter to normalize sparse inputs. Instead, Jaritz et al.\cite{jaritz2018sparse}compared different architectures and thought that using effective masks would degrade performance because masks in earlier layers in CNNs were saturated. However, this effect can be perfectly avoided by treating the binary validity masks as continuous confidence fields describing the reliability of the data proposed by Eldesokey et al.\cite{eldesokey2019confidence}. In addition, this supports confidence propagation, which helps track the reliability of data throughout the processing pipeline. 
\subsection{Guided Depth Upsampling}
Due to the sub-optimal performance of unguided methods on the edges, some recent methods urge the use of guidance from auxiliary data, such as RGB images or surface normals. Traditional methods mainly rely on local filtering techniques such as joint bilateral filtering and global optimization techiniques such as Markov random fields. 
Tomasi et al.\cite{tomasi1998bilateral}proposed bilateral filtering which produces no phantom colors along edges in color images, and reduces phantom colors where they appear in the original image.
Diebel et al.\cite{diebel2006application}exploited the fact that discontinuities in range and coloring tend to co-align to generate high-resolution, low-noise range images by integrating regular camera images into the range data.
By relying on the assumption that pixels with similar colors are likely to belong to the same surface, Favaro et al.\cite{favaro2010recovering}proposed a novel scheme to recover depth maps containing thin structures based on nonlocal-means filtering regularization. Inspired by \cite{favaro2010recovering},  the framework in \cite{park2011high}extended this regularization with an additional edge weighting scheme based on several image features based on the additional high-resolution RGB input  in order to maintain fine detail and structure. 
These methods are able to deal with both regularly and irregularly sampled data, but they use hand-crafted features that limit their performance. 

In recent years, researchers have come up with various deep learning methods. Ma et al.\cite{ma2019self}adopted an early fusion scheme, combining sparse depth input with corresponding RGB images, and achieved good results. On the other hand, Jaritz et al.\cite{jaritz2018sparse}argued that late fusion performed better on the architecture they proposed, which was also demonstrated in\cite{hua2018normalized}. Wirges et al.\cite{wirges2017guided}used a combination of RGB images and surface normals to guide the process of depth upsampling. Konno et al.\cite{konno2015intensity}used residual interpolation to combine low-resolution depth maps with high-resolution RGB images to generate high-resolution depth maps. Based on\cite{uhrig2017sparsity}, Huang et al.\cite{huang2018hms}utilized the sparsity-invariant layer to design a sparsity-invariant multi-scale encoder-decoder network for sparse depth completion with RGB guidance. Zhang et al.\cite{zhang2018deep}proposed to predict surface normal and occlusion boundary from a deep network and further utilize them to help depth completion in indoor scenes. Qiu et al.\cite{qiu2019deeplidar} extended a similar surface normal as guidance idea to the outdoor environment and recovered dense depth from sparse LiDAR data.

The aforementioned approaches are usually based on the assumption that depth edges consistently appear at the edges of the RGB images.  
However, although there is a close connections between the depth map and the color image, the discontinuity (especially the texture) of the image is not always consistent with the depth map.
Using the single color image as guidance, they ignore the structural difference between the depth and the guidance color image, and much irrelevant information such as textures on the color image will mislead upsampling task on the depth image.
To address these problems, in this study, we implement depth edge map guided convolution as a layer in CNNs for sparse inputs inspired by \cite{ye2018depth} and \cite{hua2018normalized}. Benefited from the guidance of depth edge information, our model gains better performance than other CNNs.

\section{The Proposed Method}
\begin{figure*}[h]
	\centering
	\includegraphics[width=0.9\textwidth]{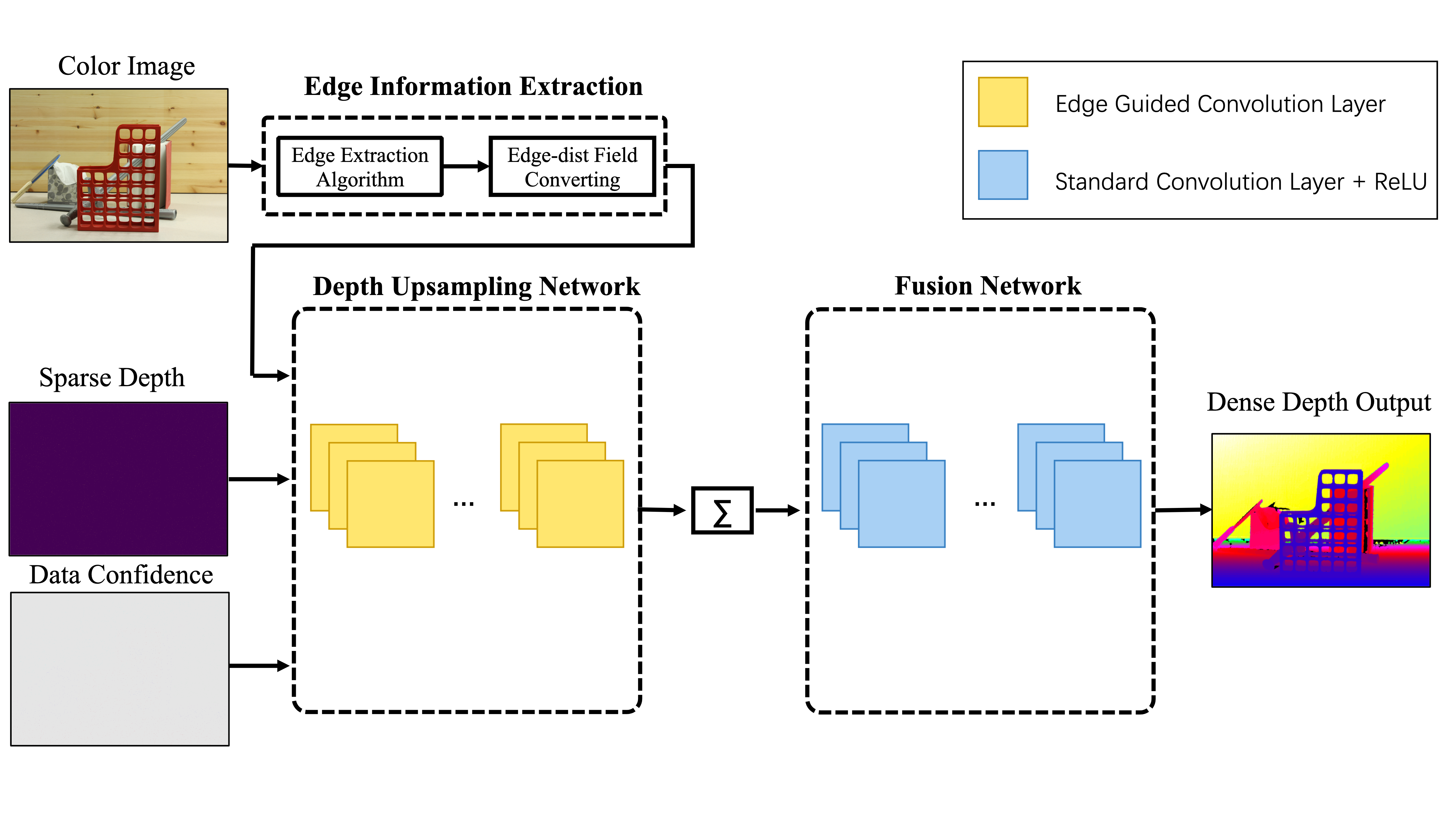}
	\caption{Our depth upsampling pipeline on an example image from the Middlebury dataset\cite{scharstein2002taxonomy,scharstein2014high}.}
	\label{EGCNN}
\end{figure*}

We propose a edge guided convolutional neural network (EGCNN) to perform sparse depth upsampling. Figure \ref{EGCNN} illustrates the entire architecture. It consists of three major components: edge information extraction component, depth upsampling  subnetwork, and depth fusion subnetwork.

The input to the pipeline is a very sparse projected LiDAR point cloud, an input confidence map which has zeros at missing pixels and ones otherwise, and a corresponding high-resolution RGB image. 
The color image is fed to an edge information extraction component to extract depth edge information. Afterwards, the continuous output depth edge map from the edge information extraction component, the sparse point cloud input and the input confidence are fed to a depth upsampling network, where we replace the standard convolutional layer by our modified edge guided convolutional layer. The outputs from the upsampling network are weighted and fed to a fusion network which produces the final dense depth map. 

To explain our edge guided convolutional network, we first elaborate the design of the edge information extraction component in subsection 3.1. Then we introduce a novel convolution layer(EGCL) in subsection 3.2. In the next, we explain how EGCL can be used in a common convolution network, which is our depth upsampling network, in subsection 3.3. Finally, we  elaborate architecture details about fusion network in subsection 3.4.




\subsection{Edge Information Extraction Component}
\begin{figure}[h]
	\centering
	\includegraphics[width=0.9\textwidth]{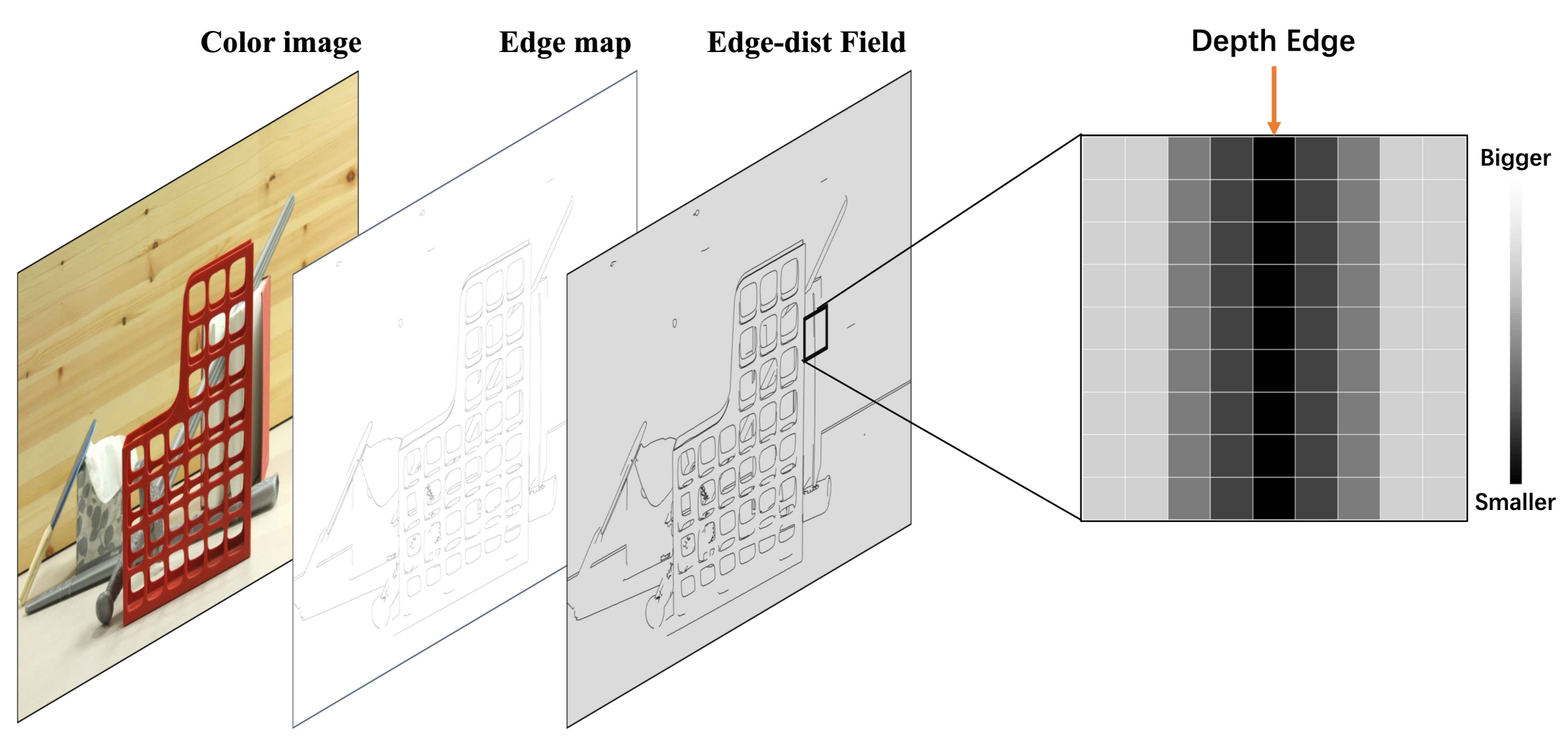}
	\caption{Process of obtaining continuous edge-dist field}
	\label{Procedure_of_Continuous Edge-dist Fields}
\end{figure}

As we observe, edges information is of especially importance in textureless depth map. Therefore, we design the edge information extraction component to obtain a continuous edge distance field from the color image, which guides the depth upsampling to generate sharp edges in the proposed method. As illustrated in Figure \ref{Procedure_of_Continuous Edge-dist Fields}, the generation of edge-dist fields consists of two steps: 1) generating edge maps using edge extraction component from color images; 2) converting the edge map into our edge-dist field. 

Edge extraction component is consist of multiple edge extraction nodes. Everyone of them takes color images $I$ as input and outputs different edge maps $P$. 

Depth map mainly contains smooth regions separated by small amounts of edges. Around the depth edge, there are always sudden change of depth value. Thus, we design a continuous edge distance field used to prevent the depth value from crossing the depth edge to facilitate upsampling. 
In order to make the depth values near the edge have less influence on the other side of the depth edge, we set smaller weights near that. For the edge map $P$, given a location $(i,j)$  which point $P_{i,j}$ is  a depth edge, the rule of the edge-dist field $E$ is defined as:
\begin{equation}
\mathrm{E}_{i, j} \leqq \mathrm{E}_{i \pm 1, j} \leqq \mathrm{E}_{i \pm 2, j} \leqq \ldots \leqq  \mathrm{E}_{i \pm k-1, j} \leqq \mathrm{E}_{i \pm k, j} \leqq \mathrm{E}_{max}
\label{Rule1}
\end{equation}
where $E_{i,j}$ is the edge-dist value of the depth edge point. As is shown in Figure \ref{Procedure_of_Continuous Edge-dist Fields}, the minima in an edge-dist field are on the depth edges. From the depth edges, the value increases gradually to the maximum value $E_{max}$ in the edge-dist field.
As the weight of data confidence, the $E_{i,j}$ and $E_{max}$ span:
\begin{equation}
\mathrm{E}_{i, j},\mathrm{E}_{max} \in[0,1]
\label{Rule2}
\end{equation}
Please note that the EGCL evaluates to a normalized convolution layer\cite{eldesokey2019confidence} when values $E$ of the edge-dist field are all $1$.

\subsection{Edge Guided Convolutional Layer}
\begin{figure}[h]
	\centering
	\includegraphics[width=0.9\textwidth]{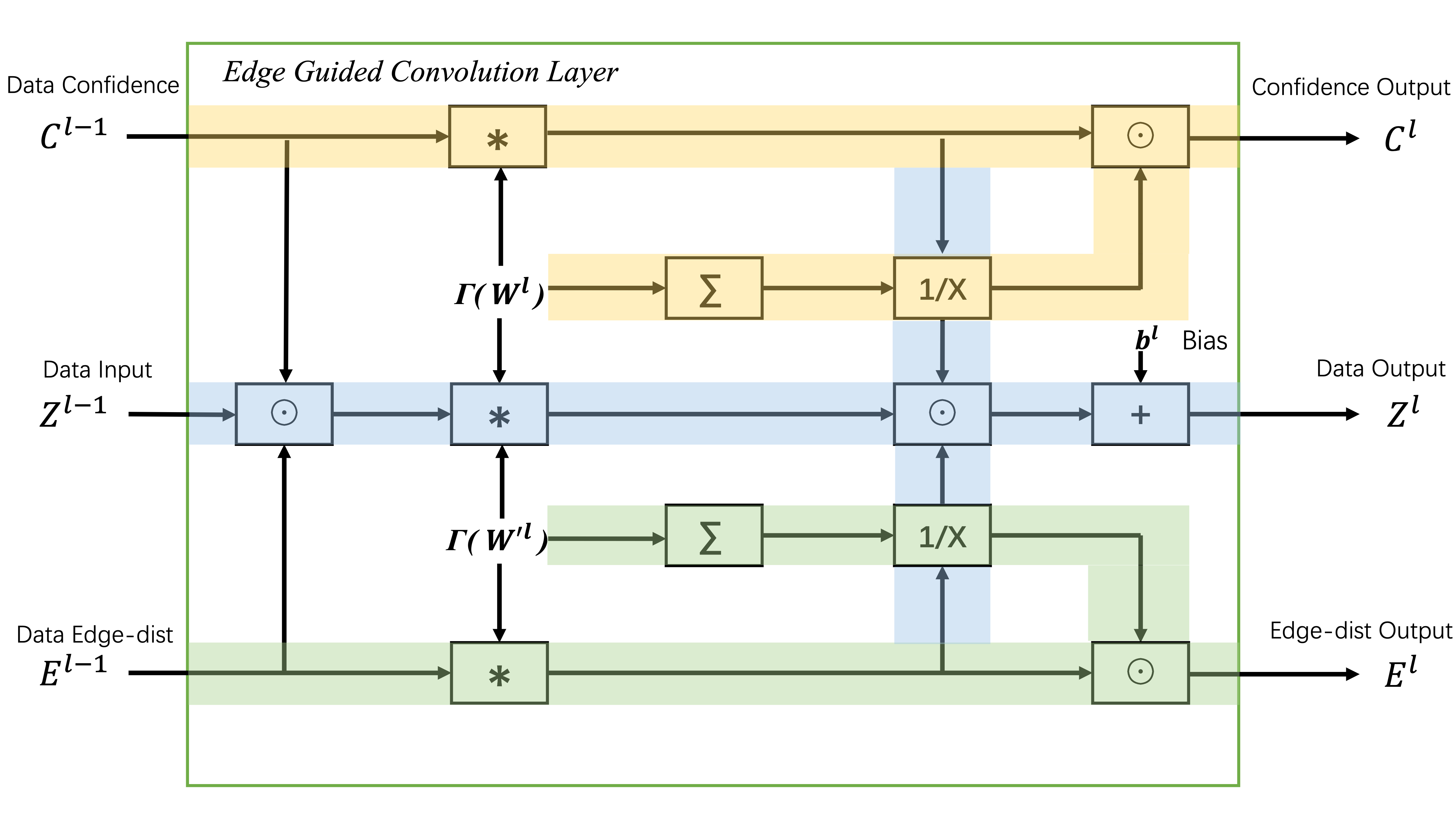}
	\caption{An illustration of the Edge Guided Convolutional layer that takes in three inputs: data, confidence and edge-dist. The Edge Guided Convolutional layer outputs a data term, a confidence term and an edge-dist term. Convolution is denoted as $\ast$, the Hadamard product (point-wise) as $\odot$, summation as $\sum$ , and point-wise inverse as 1/x.}
	\label{Edge Guided CNN}
\end{figure}

Based on the edge-dist field defined above, we propose a novel convolution layer, the edge guided convolutional layer, which using data confidence to dealing sparse inputs and continuous edge-dist fields as the guidance of depth filling. The edge guided convolutional layer can replace standard convolutional layer in the CNN framework. An illustration of the edge guided convolutional layer is shown in Figure \ref{Edge Guided CNN} .

First, the layer receives three inputs simultaneously, the sparse data, its confidence and edge-dist field. The forward pass is then modified according to \ref{Forward Pass for Edge Guided Convolution}, which is shown in Figure \ref{Edge Guided CNN} as  the blue area:
\begin{equation}
\mathbf{Z}_{i, j}^{l}=\frac{\sum_{m, n} \mathbf{Z}_{i+m, j+n}^{l-1} \mathbf{C}_{i+m,  j+n}^{l-1} \mathbf{E}_{i+m, j+n}^{l-1}\Gamma\left(\mathbf{W}_{m, n}^{l}\right)}{\sum_{m, n} \mathbf{C}_{i+m, j+n}^{l-1}\mathbf{E}_{i+m, j+n}^{l-1} \Gamma\left(\mathbf{W}_{m, n}^{l}\right)+\epsilon}+\mathbf{b}^{l}
\label{Forward Pass for Edge Guided Convolution}
\end{equation}
where $Z^l_{i,j}$ is the output of the $l^{th}$ layer at locations $i,j$ depending on the weight elements $W^l_{m,n}$, $Z^{l-1}$is the output data from the previous layer, $C^{l-1}$is the output confidence from the previous while $E^{l-1}$is the output edge-dist from the previous. $b^{l}$ is the bias and $\epsilon$ is a constant to prevent division by zero. Please note that this is formally a correlation, as it is a common notation in CNNs.

Similar to the method proposed in\cite{eldesokey2018propagating}, 
our confidence output measure is shown in Figure \ref{Edge Guided CNN} as the yellow area, which is defined as:
\begin{equation}
\mathbf{C}_{i, j}^{l}=\frac{\sum_{m, n} \mathbf{C}_{i+m, j+n}^{l-1} \Gamma\left(\mathbf{W}_{m, n}^{l}\right)+\epsilon}{\sum_{m, n} \Gamma\left(\mathbf{W}_{m, n}^{l}\right)}
\label{Propagating the Confidence}
\end{equation}
As described in \cite{jaritz2018sparse}, using effective masks would degrade performance because masks in earlier layers in CNNs were saturated, which affects several methods in \cite{uhrig2017sparsity,liu2018image,hua2018normalized}. However, \ref{Propagating the Confidence} allows propagating confidence between CNN layers without facing the above "validity mask saturation" problem.

To allow propagate edge-dist between CNN layers, as the same as what is illustrated in the yellow area of Figure \ref{Edge Guided CNN}, we define the edge-dist output measure as:
\begin{equation}
\mathbf{E}_{i, j}^{l}=\frac{\sum_{m, n} \mathbf{E}_{i+m, j+n}^{l-1} \Gamma\left(\mathbf{{W}'}_{m, n}^{l}\right)+\epsilon}{\sum_{m, n} \Gamma\left(\mathbf{{W}'}_{m, n}^{l}\right)}
\end{equation}
where  $E^{l-1}$is the output edge-dist from the previous layer, and ${{W}'}^l_{m,n}$ is the specified weight elements, different from $W^l_{m,n}$ which is utilized in \ref{Propagating the Confidence}. In order to keep the consistency of the position of the edge information, ${{W}'}^l_{m,n}$ is des igned as a filtering kernel which the intermediate element is 1 and 0 otherwise. This is expressed as:
\begin{equation}
\mathbf{{W}'}_{i, j}^{l}=\left\{\begin{array}{ll}{1,} & {\text { if }  i=\lfloor m/2 \rfloor , j=\lfloor n/2 \rfloor} \\ {0,} & {\text { otherwise }}\end{array}\right.
\end{equation}

\subsection{Depth Upsampling Network}
\begin{figure}[h]
	\centering
	\includegraphics[width=0.9\textwidth]{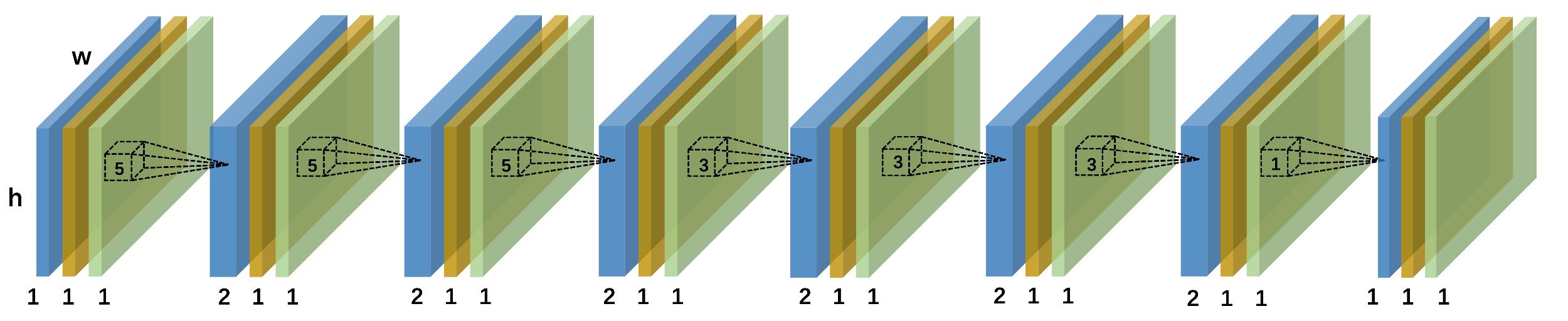}
	\caption{Our proposed depth upsampling network that utilizes Edge Guided Convolution layers. The input to our network is a sparse depth map (blue), a continuous confidence map (red) and a continuous edge-dist map (green). It passes through several Edge Guided Convolution layers (dashed) with decreasing kernel sizes from 5$\times$5 to 3$\times$3. }
	\label{Depth Upsampling Network}
\end{figure}

The depth upsampling  subnetwork is a common seven-layered convolutional neural network, but we replace all stand convolutional layers by our edge guided convolutional layers. The EGCLs in the depth upsampling subnetwork play two roles: on one hand, they are able to extract features directly from an irregularly sampled sparse depth map; on the other hand, the EGCLs also perform interpolation or upsampling for sparse data so that the produced feature maps are dense. The complete architecture of our network is illustrated in Figure \ref{Depth Upsampling Network}. 

\subsection{Fusion Network}
\begin{figure*}[!h]
	\centering
	\includegraphics[width=0.9\textwidth]{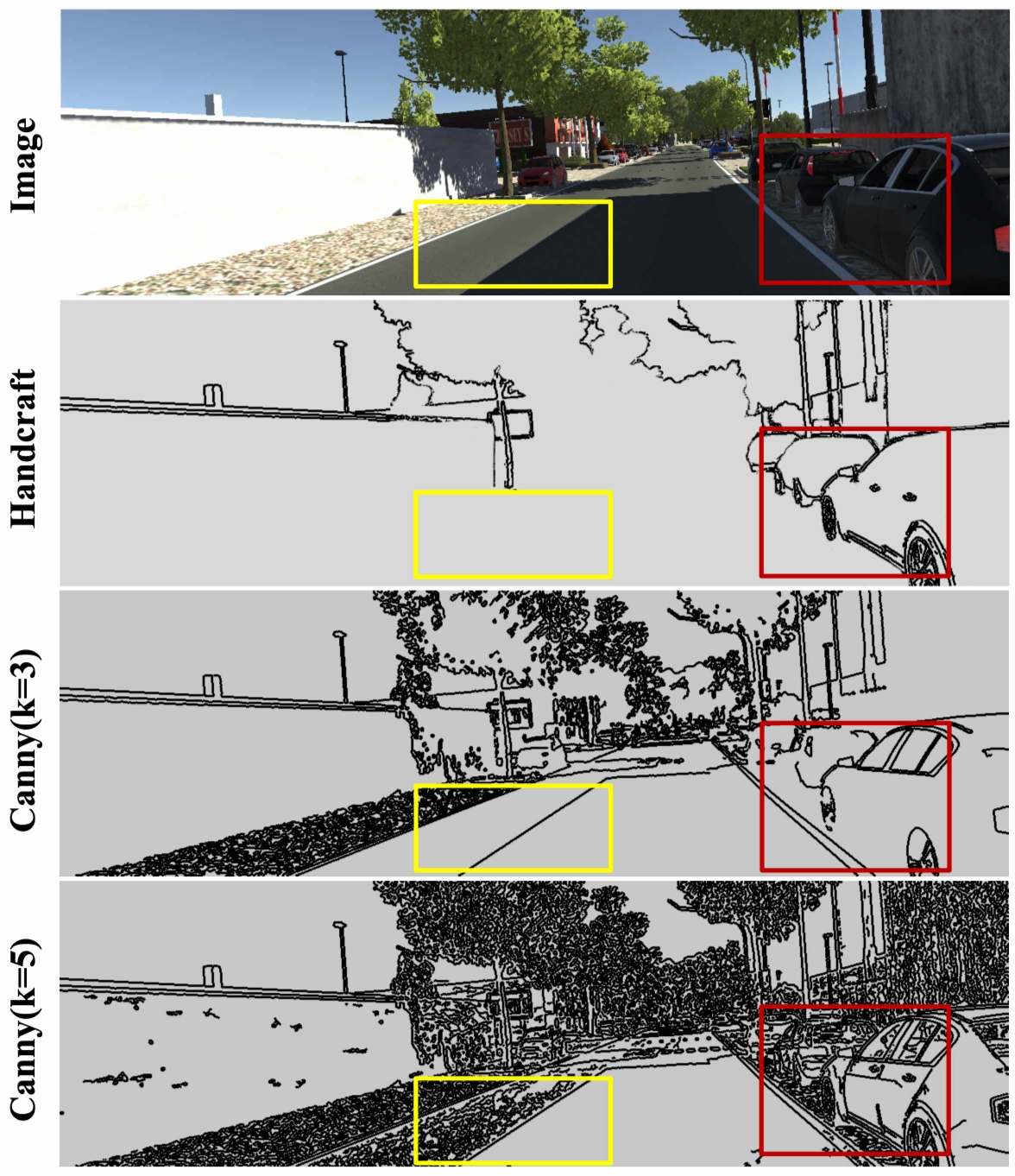}
	\caption{Qualitative comparison for different edge maps. Color image is shown in the top row. Our handcrafted depth map is 'Handcraft'. The edge map which generated by the Canny operator (the upper threshold is 150, the lower threshold is 50, kernel size is 3) is 'Canny(k=3)', while 'Canny(k=5)' is the edge map which generated by the Canny operator (the upper threshold is 150, the lower threshold is 50, kernel size is 5).}
	\label{Qualitative comparison for different edge maps}
\end{figure*}

Different edge extraction algorithms have their own limits and deficiencies, so do the dense depth images which guided by their  different edge maps. The depth fusion subnetwork takes the weighted dense depth images from depth upsampling network which guided by different edge maps as the input. As is shown in Figure \ref{Qualitative comparison for different edge maps}, the edge maps generated by the Canny operator (the upper threshold is 150, the lower threshold is 50, kernel size is 3) lose some depth edge details such as outline of cars compared with our handcrafted depth map(red rectangle in Figure \ref{Qualitative comparison for different edge maps}). Although it contains much more details, the maps generated by the Canny operator (the upper threshold is 150, the lower threshold is 50, kernel size is 5) suffer from too much misleading edges made by textures on the object surface (yellow rectangle in Figure \ref{Qualitative comparison for different edge maps}). These local dense edges will make the edge maps meaningless locally.
So we design the fusion subnetwork. It can improve  upsampling ability by combining multiple results guided by different edge maps into a single new result. It makes us to combine the advantages of different algorithms and achieve better performance. 

{\LARGE }
\section{Experiments}
To demonstrate the capabilities of our proposed approach, we evaluate our proposed networks on the Virtual KITTI \cite{Gaidon:Virtual:CVPR2016} dataset and the Middlebury \cite{scharstein2002taxonomy,scharstein2014high}dataset for the task of sparse depth upsampling.
\subsection{Datasets}
Virtual KITTI is a photo-realistic synthetic video dataset designed to learn and evaluate computer vision models for several video understanding tasks.Virtual KITTI contains 50 high-resolution monocular videos (21,260 frames) generated from five different virtual worlds in urban settings under different imaging and weather conditions.These worlds were created using the Unity game engine and a novel real-to-virtual cloning method. These photo-realistic synthetic videos are at the pixel level with depth labels. 

The Middlebury is a high-resolution stereo datasets of static indoor scenes with highly accurate ground-truth disparities. Please note that all the  disparities images have been converted to depth images. 
\subsection{Experimental Setup}
Both datasets provide us with dense depth maps and aligned color images. For experiments we synthetically generate sparse depth maps with different sampling rates by randomly sampling a percentage of points from the provided dense depth maps.

\subsection{Evaluation Metrics}
For the Virtual KITTI dataset, we compute the Mean Absolute Error (MAE) and the Root Mean Square Error (RMSE) on the depth values. The MAE is an unbiased error metric, takeing an average of the error across the whole image and it is defined as:
\begin{equation}
\operatorname{MAE(Z, T)}=\frac{1}{M N}\left[\sum_{i=0}^{N} \sum_{j=0}^{M}|Z(i, j)-T(i, j)|\right],
\end{equation}
where $Z(i,j)$ is the data output from our method and $T(i,j)$is the data ground truth. The RMSE penalizes outliers and it is defined as:
\begin{equation}
\operatorname{RMSE(Z,T)}=\frac{1}{M N}\left[\sum_{i=0}^{N} \sum_{j=0}^{M}|Z(i, j)-T(i, j)|^{2}\right]^{1 / 2}.
\end{equation}
Additionally, we also use iMAE and iRMSE, which are calculated on the disparity instead of the depth. The 'i' indicates that disparity is proportional to the inverse of
depth.

For the Middlebury dataset, we compute the RMSE, the MAE, and the inliers ratio $\delta_{i}$ as descriped in \cite{eigen2014depth}, which means the percentage of predicted pixels where the relative error is less a threshold $i$. Specifically, $i$ is chosen as 1.25,$1.25^{2}$ and $1.25^{3}$ separately for evaluation. Here, a higher $i$ indicates a softer constraint and a higher $\delta_{i}$ represents a better prediction. $\delta_{n}$ is defined as:
\begin{equation}
\delta_{n}=\frac{n\left(\left\{Z: \max \left\{\frac{Z}{T}, \frac{T}{Z}\right\}<1.25^{n}\right\}\right)}{n\left(\left\{T\right\}\right)}
\end{equation}
with n(·) as the cardinality of a set. 
\begin{table}[ht]
	\caption{Quantitative results on the Virtual KITTI dataset with the models trained at various sampling rates and tested at corresponding rates.(The best one is  highlighted in bold.)}  
	\centering
	\label{table1}  
	\begin{tabular}{c|c|cccc}
		\hline
		method       & \multicolumn{1}{l|}{Samples} & \multicolumn{1}{l}{RMSE} & \multicolumn{1}{r}{MAE} & \multicolumn{1}{l}{iRMSE} & \multicolumn{1}{l}{iMAE} \\ \hline
		Sparse   & \multirow{3}{*}{5\%}         & 15.931                   & 3.271                   & 0.075                     & 0.035                    \\
		Normal        &                              & 16.82                    & 2.489                   & 0.076                     & 0.012                    \\
		Edge(ours) &                              & \textbf{15.680}          & \textbf{2.340}           & \textbf{0.070}            & \textbf{0.007}           \\ \hline
		Sparse   & \multirow{3}{*}{0.8\%}       & 21.323                   & 4.427                   & 0.086                     & 0.037                    \\
		Normal        &                              & 23.046                   & 3.934                   & 0.089                     & 0.019                    \\
		Edge(ours) &                              & \textbf{21.063}          & \textbf{3.591}          & \textbf{0.071}            & \textbf{0.017}           \\ \hline
		Sparse   & \multirow{3}{*}{0.2\%}       & 26.722                   & 6.056                   & 0.092                     & 0.039                    \\
		Normal        &                              & 28.824                   & 5.581                   & 0.097                     & 0.031                    \\
		Edge(ours) &                              & \textbf{26.584}          & \textbf{5.111}          & \textbf{0.079}            & \textbf{0.025}           \\ \hline
	\end{tabular}
\end{table}

\begin{table*}[ht]
	\centering
	\caption{Quantitative results on the Middlebury dataset with the models trained at various sampling rates and tested at corresponding rates.(The best one is  highlighted in bold.)}
	\label{table2} 
	\begin{tabular}{l|c|cc|ccc}
		\hline
		\multicolumn{1}{c|}{\multirow{2}{*}{method}} & \multicolumn{1}{l|}{\multirow{2}{*}{samples}} & \multicolumn{2}{l|}{Lower the Better}               & \multicolumn{3}{l|}{Higher the Better}                                   \\ \cline{3-7} 
		\multicolumn{1}{c|}{}                        & \multicolumn{1}{l|}{}                         & \multicolumn{1}{l}{RMSE} & \multicolumn{1}{l|}{MAE} & \multicolumn{1}{l}{$\delta_{1}$} & \multicolumn{1}{l}{$\delta_{2}$} & \multicolumn{1}{l}{$\delta_{3}$} \\ \hline
		Sparse                                   & \multirow{3}{*}{5\%}                          & 1.623                    & 0.339                    & 0.922                  & 0.952                  & 0.972                  \\
		Normal                                        &                                               & 1.672                    & 0.344                    & 0.919                  & 0.951                  & 0.972                  \\
		Edge(ours)                        &                                               & \textbf{1.573}           & \textbf{0.282}           & \textbf{0.927}         & \textbf{0.959}         & \textbf{0.981}         \\ \hline
		Sparse                                   & \multirow{3}{*}{0.8\%}                        & 2.702                    & 0.616                    & 0.861                  & 0.913                  & 0.954                  \\
		Normal                                        &                                               & 2.814                    & 0.604                    & 0.863                  & 0.916                  & 0.958                  \\
		Edge(ours)                        &                                               & \textbf{2.676}           & \textbf{0.548}           & \textbf{0.868}         & \textbf{0.921}         & \textbf{0.964}         \\ \hline
		Sparse                                   & \multirow{3}{*}{0.2\%}                        & 4.245                    & 1.083                    & 0.783                  & 0.856                  & 0.922                  \\
		Normal                                        &                                               & 4.309                    & 1.057                    & 0.788                  & 0.862                  & 0.930                  \\
		Edge(ours)                        &                                               & \textbf{4.222}           & \textbf{1.016}           & \textbf{0.790}         & \textbf{0.866}         & \textbf{0.933}         \\ \hline
	\end{tabular}
\end{table*}

\subsection{Evaluating Edge Guided Convolution Layer}
We run a number of experiments to analyze our proposed convolution layer. To investigate the effectiveness of EGCL, we compare our models with those replacing EGCLs with either normalized convolutional layers\cite{eldesokey2019confidence} or sparse convolutional layers\cite{uhrig2017sparsity}. These models are, respectively, denoted using the suffixes ‘Edge’, ‘ Norm’ and ‘ Sparse’.

In experiments, to compare the performance on sparse data whose sampling rate is not more than 5\%, we train all models  at various sampling rates and test them at corresponding rates. Experiments are conducted on both datasets, from which consistent phenomena can be observed. We present the quantitative results on the Virtual KITTI dataset in Table \ref{table1} and the results on the Middlebury dataset in Table \ref{table2}. The value in bold is the best. 

Our model is improved on the basis of the normalized convolutional layer to generate sharper edges so that we can get less errors in the area around depth edges. From the tables, we can conclude that our method  rank the first in the test results. Especially, at 5\% sampling density, our MAE decreased by 28.5\% and 16.8\% on the Virtual KITTI dataset and the Middlebury dataset respectively. 

\begin{figure*}[ht]
	\centering
	\includegraphics[width=0.9\textwidth]{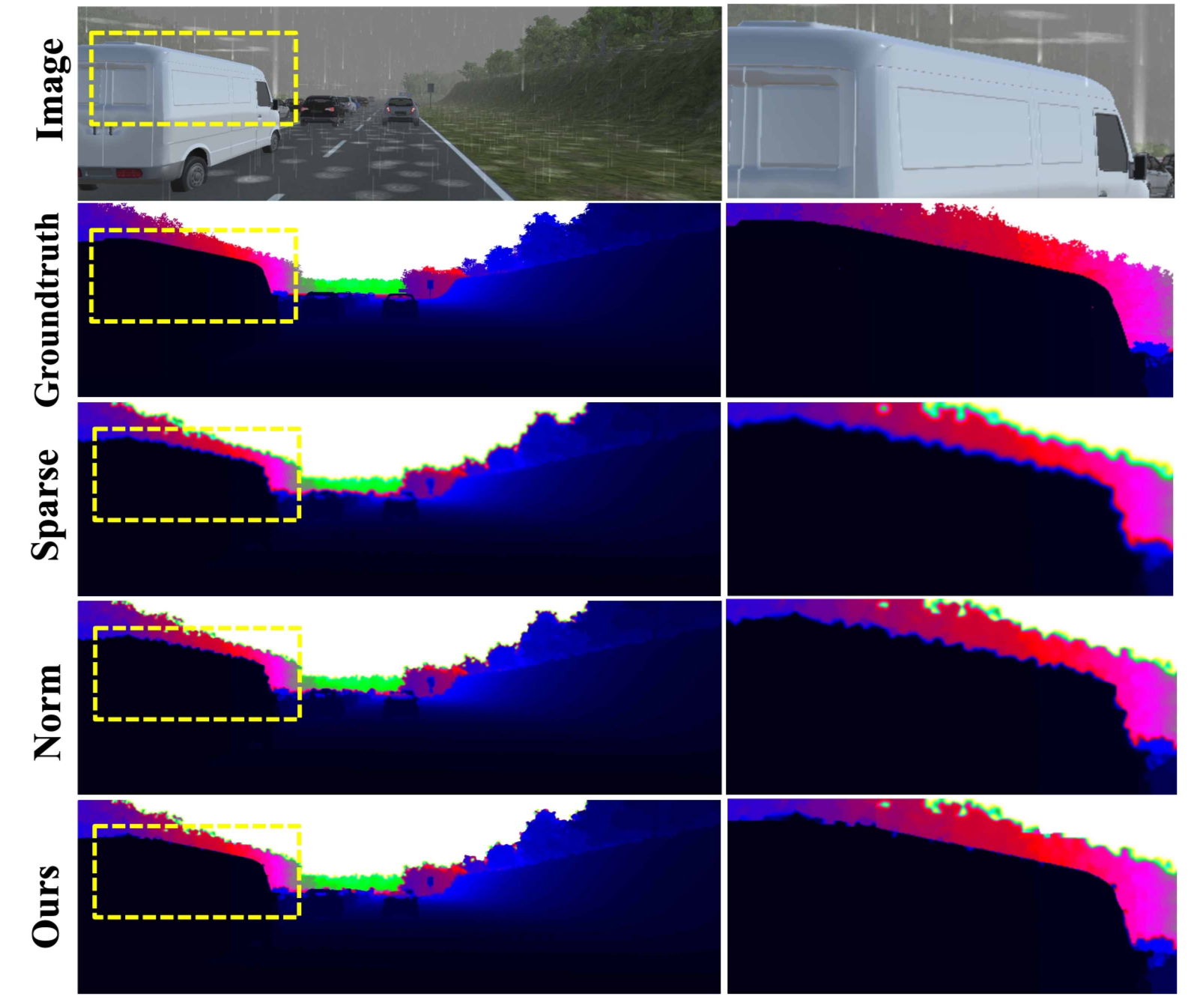}
	\caption{Qualitative comparison with state-of-the-art methods on the Virtual KITTI dataset under the 'rain' conditions. The depth images are colorized along with depth range. Our results are shown in the bottom row and compared with top-ranking methods ‘Sparse’\cite{uhrig2017sparsity}, ‘Normal’\cite{eldesokey2019confidence}. In the zoomed regions, our method preserves sharper edges.}
	\label{Qualitative comparison on the Virtual KITTI dataset}
\end{figure*}

\begin{figure*}[h]
	\centering
	\includegraphics[width=0.9\textwidth]{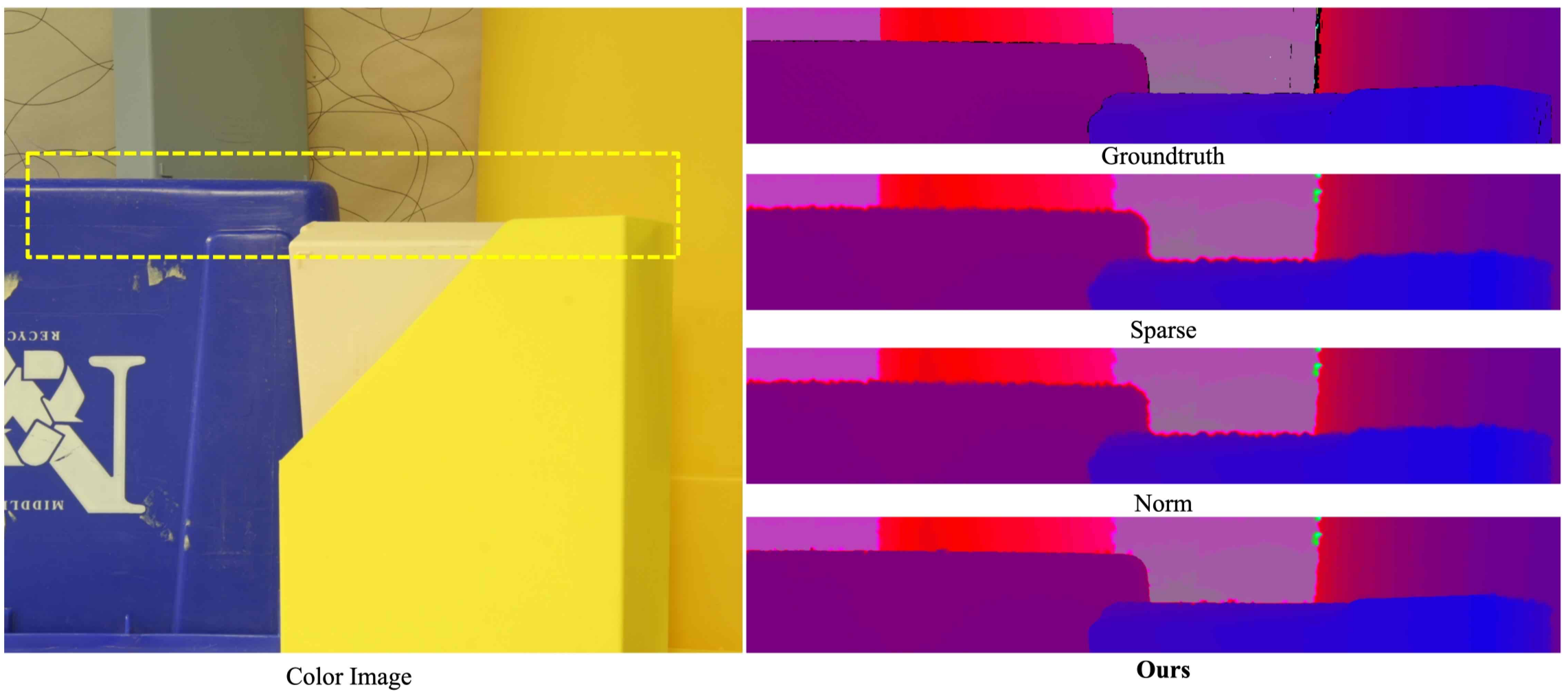}
	\caption{Qualitative comparison with state-of-the-art methods on the Middlebury dataset image 'Plastic'. The depth images are colorized along with depth range. Our results are shown in the bottom row and compared with top-ranking methods ‘Sparse’\cite{uhrig2017sparsity}, ‘Normal’\cite{eldesokey2019confidence}. In the zoomed regions, our method recovers better 3D details.}
	\label{Qualitative comparison on the Middlebury dataset}
\end{figure*}

Figure \ref{Qualitative comparison on the Virtual KITTI dataset}
shows some qualitative results on the Virtual KITTI dataset 
while Figure \ref{Qualitative comparison on the Middlebury dataset}
shows some on the Middlebury dataset. The predictions from 'Sparse' and 'Normal' are very blurry especially along depth edges. However,with the help of guidance by edge-dist field, our methods EGCL produce remarkably better reconstructions of the dense map, in particular with respect to edges sharpness and the level of details.


\subsection{Evaluating Edge Guided CNNs}
To investigate the effectiveness of EGCNN, we then compare our full model EGCNN-Full to other ablation network structure which do not have fusion subnetwork, denoted as EGCNN-Ablations. Among them, the ablation network which use Canny(the upper threshold is 150, the lower threshold is 50,kernel size is 3) as the edge extraction algorithm is EGCNN-Ablation(k=3), while  EGCNN-Ablation(k=5) uses Canny(the upper threshold is 150, the lower threshold is 50, kernel size is 5). 
Table \ref{table3} and  \ref{table4} report the experimental results at five sampling rates (5\%, 0.8\%, 0.2\%) on both datasets. As the same as the above, experiments are trained and tested at the same rate. 

From Table \ref{table3} and \ref{table4} we get the following observations: (1)In most cases, EGCNN-Ablation(k=5) achieve better results than EGCNN-Ablation(k=3). As we can see in Figure \ref{Qualitative comparison for different edge maps}, although there are more useful details in 'Canny(k=5)', it do contains much more wrong edge information even very dense misleading edge lines. Quantitative experimental results show that wrong edge information can hardly make the performance of EGCL wrose but correct edge information do improve it. (2) With the help of fusion subnetwork, EGCNN-Full can combine the advantages of different edge extraction algorithms and achieve further better performance. Both tables show that our proposed model outperforms both ablation models in most cases, especially in RMSE.

\begin{table*}[ht]
	\centering
	\caption{Quantitative results on the Vkitti dataset with the models trained at various sampling rates and tested at corresponding rates.(The best one is  highlighted in bold.)}  
	\label{table3}  
	\begin{tabular}{c|c|cccc}
		\hline
		method             & \multicolumn{1}{l|}{Samples} & \multicolumn{1}{l}{RMSE} & \multicolumn{1}{r}{MAE} & \multicolumn{1}{l}{iRMSE} & \multicolumn{1}{l}{iMAE} \\ \hline
		EGCNN-Ablation(k3) & \multirow{3}{*}{5\%}         & 15.680                   & 2.340                   & 0.070                     & 0.007                    \\
		EGCNN-Ablation(k5) &                              & 15.674                   & 2.338                   & 0.069                     & 0.007                    \\
		EGCNN-Full         &                              & \textbf{15.574}          & \textbf{2.336}          & \textbf{0.068}            & \textbf{0.006}           \\ \hline
		EGCNN-Ablation(k3) & \multirow{3}{*}{0.8\%}       & 21.063                   & 3.591                   & 0.071                     & \textbf{0.017}           \\
		EGCNN-Ablation(k5) &                              & 21.056                   & 3.596                   & 0.070                     & \textbf{0.017}           \\
		EGCNN-Full         &                              & \textbf{20.921}          & \textbf{3.590}          & \textbf{0.069}            & \textbf{0.017}           \\ \hline
		EGCNN-Ablation(k3) & \multirow{3}{*}{0.2\%}       & 26.584                   & 5.111                   & 0.079                     & \textbf{0.025}           \\
		EGCNN-Ablation(k5) &                              & 26.577                   & 5.111                   & 0.077                     & 0.026                    \\
		EGCNN-Full         &                              & \textbf{26.536}          & \textbf{5.106}          & \textbf{0.076}            & \textbf{0.025}           \\ \hline
	\end{tabular}
\end{table*}

\begin{table*}[ht]
	\centering
	\caption{Quantitative results on the Middleburry dataset with the models trained at various sampling rates and tested at corresponding rates.(The best one is  highlighted in bold.)}  
	\label{table4}  
	\begin{tabular}{l|c|cc|ccc}
		\hline
		\multirow{2}{*}{method}         & \multicolumn{1}{l|}{\multirow{2}{*}{samples}} & \multicolumn{2}{l|}{Lower the Better}               & \multicolumn{3}{l|}{Higher the Better}                                   \\ \cline{3-7} 
		& \multicolumn{1}{l|}{}                         & \multicolumn{1}{l}{RMSE} & \multicolumn{1}{r|}{MAE} & \multicolumn{1}{l}{{$\delta_{1}$} } & \multicolumn{1}{l}{{$\delta_{2}$} } & \multicolumn{1}{l}{{$\delta_{3}$} } \\ \hline
		EGCNN-Ablation(k3)              & \multirow{3}{*}{5\%}                          & 1.573                    & 0.282                    & 0.927                  & 0.959                  & 0.981                  \\
		EGCNN-Ablation(k5)              &                                               & 1.571                    & 0.281                    & 0.933                  & \textbf{0.977}         & \textbf{1.000}         \\
		\multicolumn{1}{c|}{EGCNN-Full} &                                               & \textbf{1.559}           & \textbf{0.280}           & \textbf{0.942}         & 0.974                  & 0.996                  \\ \hline
		EGCNN-Ablation(k3)              & \multirow{3}{*}{0.8\%}                        & 2.676                    & \textbf{0.548}           & 0.868                  & \textbf{0.921}         & \textbf{0.964}         \\
		EGCNN-Ablation(k5)              &                                               & 2.669                    & 0.549                    & 0.860                  & 0.913                  & 0.955                  \\
		\multicolumn{1}{c|}{EGCNN-Full} &                                               & \textbf{2.655}           & \textbf{0.548}           & \textbf{0.869}         & 0.911                  & 0.957                  \\ \hline
		EGCNN-Ablation(k3)              & \multirow{3}{*}{0.2\%}                        & 4.222                    & 1.016                    & 0.790                  & 0.866                  & 0.933                  \\
		EGCNN-Ablation(k5)              &                                               & 4.208                    & \textbf{1.014}           & 0.781                  & 0.861                  & 0.934                  \\
		\multicolumn{1}{c|}{EGCNN-Full} &                                               & \textbf{4.190}           & \textbf{1.014}           & \textbf{0.803}         & \textbf{0.881}         & \textbf{0.942}         \\ \hline
	\end{tabular}
\end{table*}

\section{Conclusion}
We propose a guided convolutional layer to recover dense depth from sparse and irregular depth image with an depth edge image as guidance. Our novel guided network can prevent the depth value from crossing the depth edge to facilitate upsampling.

We further design a convolution network based on proposed convolutional layer to combine the advantages of different algorithms and achieve better performance.

Extensive experiments and ablation studies verify the superior performance of our guided convolutional layer and the effectiveness of the convolution network on sparse depth upsampling. Our method not only shows strong results on both indoor and outdoor, virtual and realistic scenes, but also presents strong generalization capability under different point densities, various lighting and weather conditions. While this paper specifically focuses on the problem of sparse depth upsampling, we believe that other tasks in computer vision involving image edges can also benefit from the design of our edge guided convolution layer and the convolution network in this paper.

\section*{Acknowledgements}
This work has been partially funded by the Chongqing Research Program of Basic Research and Frontier Technology grant No.cstc2019jcyj-msxmX0033, the National Natural Science Foundation of China grant No.61701051, the Fundamental Research Funds for the Central Universities grant No.2019CDYGYB012.
\bibliographystyle{unsrt} 
\bibliography{mybibfile}

\begin{thebibliography}{10}

\bibitem{Velodyne2018}
Velodyne.
\newblock \url{http://velodynelidar.com/hdl-64e.html}, 2018.

\bibitem{yang2019std}
Zetong Yang, Yanan Sun, Shu Liu, Xiaoyong Shen, and Jiaya Jia.
\newblock Std: Sparse-to-dense 3d object detector for point cloud.
\newblock In {\em Proceedings of the IEEE International Conference on Computer
  Vision}, pages 1951--1960, 2019.

\bibitem{liang2019multi}
Ming Liang, Bin Yang, Yun Chen, Rui Hu, and Raquel Urtasun.
\newblock Multi-task multi-sensor fusion for 3d object detection.
\newblock In {\em Proceedings of the IEEE Conference on Computer Vision and
  Pattern Recognition}, pages 7345--7353, 2019.

\bibitem{du2018general}
Xinxin Du, Marcelo~H Ang, Sertac Karaman, and Daniela Rus.
\newblock A general pipeline for 3d detection of vehicles.
\newblock In {\em 2018 IEEE International Conference on Robotics and Automation
  (ICRA)}, pages 3194--3200. IEEE, 2018.

\bibitem{qi2018frustum}
Charles~R Qi, Wei Liu, Chenxia Wu, Hao Su, and Leonidas~J Guibas.
\newblock Frustum pointnets for 3d object detection from rgb-d data.
\newblock In {\em Proceedings of the IEEE Conference on Computer Vision and
  Pattern Recognition}, pages 918--927, 2018.

\bibitem{caltagirone2019lidar}
Luca Caltagirone, Mauro Bellone, Lennart Svensson, and Mattias Wahde.
\newblock Lidar--camera fusion for road detection using fully convolutional
  neural networks.
\newblock {\em Robotics and Autonomous Systems}, 111:125--131, 2019.

\bibitem{chen2019progressive}
Zhe Chen, Jing Zhang, and Dacheng Tao.
\newblock Progressive lidar adaptation for road detection.
\newblock {\em IEEE/CAA Journal of Automatica Sinica}, 6(3):693--702, 2019.

\bibitem{riegler2016atgv}
Gernot Riegler, Matthias R{\"u}ther, and Horst Bischof.
\newblock Atgv-net: Accurate depth super-resolution.
\newblock In {\em European conference on computer vision}, pages 268--284.
  Springer, 2016.

\bibitem{uhrig2017sparsity}
Jonas Uhrig, Nick Schneider, Lukas Schneider, Uwe Franke, Thomas Brox, and
  Andreas Geiger.
\newblock Sparsity invariant cnns.
\newblock In {\em 2017 International Conference on 3D Vision (3DV)}, pages
  11--20. IEEE, 2017.

\bibitem{tomasi1998bilateral}
Carlo Tomasi and Roberto Manduchi.
\newblock Bilateral filtering for gray and color images.
\newblock In {\em Sixth international conference on computer vision (IEEE Cat.
  No. 98CH36271)}, pages 839--846. IEEE, 1998.

\bibitem{petschnigg2004digital}
Georg Petschnigg, Richard Szeliski, Maneesh Agrawala, Michael Cohen, Hugues
  Hoppe, and Kentaro Toyama.
\newblock Digital photography with flash and no-flash image pairs.
\newblock {\em ACM transactions on graphics (TOG)}, 23(3):664--672, 2004.

\bibitem{diebel2006application}
James Diebel and Sebastian Thrun.
\newblock An application of markov random fields to range sensing.
\newblock In {\em Advances in neural information processing systems}, pages
  291--298, 2006.

\bibitem{park2011high}
Jaesik Park, Hyeongwoo Kim, Yu-Wing Tai, Michael~S Brown, and Inso Kweon.
\newblock High quality depth map upsampling for 3d-tof cameras.
\newblock In {\em 2011 International Conference on Computer Vision}, pages
  1623--1630. IEEE, 2011.

\bibitem{riegler2016deep}
Gernot Riegler, David Ferstl, Matthias R{\"u}ther, and Horst Bischof.
\newblock A deep primal-dual network for guided depth super-resolution.
\newblock {\em arXiv preprint arXiv:1607.08569}, 2016.

\bibitem{hui2016depth}
Tak-Wai Hui, Chen~Change Loy, and Xiaoou Tang.
\newblock Depth map super-resolution by deep multi-scale guidance.
\newblock In {\em European conference on computer vision}, pages 353--369.
  Springer, 2016.

\bibitem{li2016deep}
Yijun Li, Jia-Bin Huang, Narendra Ahuja, and Ming-Hsuan Yang.
\newblock Deep joint image filtering.
\newblock In {\em European Conference on Computer Vision}, pages 154--169.
  Springer, 2016.

\bibitem{chan2008noise}
Derek Chan, Hylke Buisman, Christian Theobalt, and Sebastian Thrun.
\newblock A noise-aware filter for real-time depth upsampling.
\newblock 2008.

\bibitem{dolson2010upsampling}
Jennifer Dolson, Jongmin Baek, Christian Plagemann, and Sebastian Thrun.
\newblock Upsampling range data in dynamic environments.
\newblock In {\em 2010 IEEE Computer Society Conference on Computer Vision and
  Pattern Recognition}, pages 1141--1148. IEEE, 2010.

\bibitem{kopf2007joint}
Johannes Kopf, Michael~F Cohen, Dani Lischinski, and Matt Uyttendaele.
\newblock Joint bilateral upsampling.
\newblock In {\em ACM Transactions on Graphics (ToG)}, volume~26, page~96. ACM,
  2007.

\bibitem{liu2013joint}
Ming-Yu Liu, Oncel Tuzel, and Yuichi Taguchi.
\newblock Joint geodesic upsampling of depth images.
\newblock In {\em Proceedings of the IEEE conference on computer vision and
  pattern recognition}, pages 169--176, 2013.

\bibitem{yang2007spatial}
Qingxiong Yang, Ruigang Yang, James Davis, and David Nist{\'e}r.
\newblock Spatial-depth super resolution for range images.
\newblock In {\em 2007 IEEE Conference on Computer Vision and Pattern
  Recognition}, pages 1--8. IEEE, 2007.

\bibitem{barron2016fast}
Jonathan~T Barron and Ben Poole.
\newblock The fast bilateral solver.
\newblock In {\em European Conference on Computer Vision}, pages 617--632.
  Springer, 2016.

\bibitem{ferstl2013image}
David Ferstl, Christian Reinbacher, Rene Ranftl, Matthias R{\"u}ther, and Horst
  Bischof.
\newblock Image guided depth upsampling using anisotropic total generalized
  variation.
\newblock In {\em Proceedings of the IEEE International Conference on Computer
  Vision}, pages 993--1000, 2013.

\bibitem{hawe2011dense}
Simon Hawe, Martin Kleinsteuber, and Klaus Diepold.
\newblock Dense disparity maps from sparse disparity measurements.
\newblock In {\em 2011 International Conference on Computer Vision}, pages
  2126--2133. IEEE, 2011.

\bibitem{schneider2016semantically}
Nick Schneider, Lukas Schneider, Peter Pinggera, Uwe Franke, Marc Pollefeys,
  and Christoph Stiller.
\newblock Semantically guided depth upsampling.
\newblock In {\em German Conference on Pattern Recognition}, pages 37--48.
  Springer, 2016.

\bibitem{song2016deep}
Xibin Song, Yuchao Dai, and Xueying Qin.
\newblock Deep depth super-resolution: Learning depth super-resolution using
  deep convolutional neural network.
\newblock In {\em Asian conference on computer vision}, pages 360--376.
  Springer, 2016.

\bibitem{eldesokey2019confidence}
Abdelrahman Eldesokey, Michael Felsberg, and Fahad~Shahbaz Khan.
\newblock Confidence propagation through cnns for guided sparse depth
  regression.
\newblock {\em IEEE transactions on pattern analysis and machine intelligence},
  2019.

\bibitem{hou1978cubic}
Hsieh Hou and H~Andrews.
\newblock Cubic splines for image interpolation and digital filtering.
\newblock {\em IEEE Transactions on acoustics, speech, and signal processing},
  26(6):508--517, 1978.

\bibitem{yang2010image}
Jianchao Yang, John Wright, Thomas~S Huang, and Yi~Ma.
\newblock Image super-resolution via sparse representation.
\newblock {\em IEEE transactions on image processing}, 19(11):2861--2873, 2010.

\bibitem{dong2015image}
Chao Dong, Chen~Change Loy, Kaiming He, and Xiaoou Tang.
\newblock Image super-resolution using deep convolutional networks.
\newblock {\em IEEE transactions on pattern analysis and machine intelligence},
  38(2):295--307, 2015.

\bibitem{kim2016accurate}
Jiwon Kim, Jung Kwon~Lee, and Kyoung Mu~Lee.
\newblock Accurate image super-resolution using very deep convolutional
  networks.
\newblock In {\em Proceedings of the IEEE conference on computer vision and
  pattern recognition}, pages 1646--1654, 2016.

\bibitem{dahl2017pixel}
Ryan Dahl, Mohammad Norouzi, and Jonathon Shlens.
\newblock Pixel recursive super resolution.
\newblock In {\em Proceedings of the IEEE International Conference on Computer
  Vision}, pages 5439--5448, 2017.

\bibitem{ma2019self}
Fangchang Ma, Guilherme~Venturelli Cavalheiro, and Sertac Karaman.
\newblock Self-supervised sparse-to-dense: Self-supervised depth completion
  from lidar and monocular camera.
\newblock In {\em 2019 International Conference on Robotics and Automation
  (ICRA)}, pages 3288--3295. IEEE, 2019.

\bibitem{chodosh2018deep}
Nathaniel Chodosh, Chaoyang Wang, and Simon Lucey.
\newblock Deep convolutional compressed sensing for lidar depth completion.
\newblock In {\em Asian Conference on Computer Vision}, pages 499--513.
  Springer, 2018.

\bibitem{hua2018normalized}
Jiashen Hua and Xiaojin Gong.
\newblock A normalized convolutional neural network for guided sparse depth
  upsampling.
\newblock In {\em IJCAI}, pages 2283--2290, 2018.

\bibitem{jaritz2018sparse}
Maximilian Jaritz, Raoul De~Charette, Emilie Wirbel, Xavier Perrotton, and
  Fawzi Nashashibi.
\newblock Sparse and dense data with cnns: Depth completion and semantic
  segmentation.
\newblock In {\em 2018 International Conference on 3D Vision (3DV)}, pages
  52--60. IEEE, 2018.

\bibitem{favaro2010recovering}
Paolo Favaro.
\newblock Recovering thin structures via nonlocal-means regularization with
  application to depth from defocus.
\newblock In {\em 2010 IEEE Computer Society Conference on Computer Vision and
  Pattern Recognition}, pages 1133--1140. IEEE, 2010.

\bibitem{wirges2017guided}
Sascha Wirges, Bj{\"o}rn Roxin, Eike Rehder, Tilman K{\"u}hner, and Martin
  Lauer.
\newblock Guided depth upsampling for precise mapping of urban environments.
\newblock In {\em 2017 IEEE Intelligent Vehicles Symposium (IV)}, pages
  1140--1145. IEEE, 2017.

\bibitem{konno2015intensity}
Yosuke Konno, Yusuke Monno, Daisuke Kiku, Masayuki Tanaka, and Masatoshi
  Okutomi.
\newblock Intensity guided depth upsampling by residual interpolation.
\newblock In {\em The Abstracts of the international conference on advanced
  mechatronics: toward evolutionary fusion of IT and mechatronics: ICAM
  2015.6}, pages 1--2. The Japan Society of Mechanical Engineers, 2015.

\bibitem{huang2018hms}
Zixuan Huang, Junming Fan, Shuai Yi, Xiaogang Wang, and Hongsheng Li.
\newblock Hms-net: Hierarchical multi-scale sparsity-invariant network for
  sparse depth completion.
\newblock {\em arXiv preprint arXiv:1808.08685}, 2018.

\bibitem{zhang2018deep}
Yinda Zhang and Thomas Funkhouser.
\newblock Deep depth completion of a single rgb-d image.
\newblock In {\em Proceedings of the IEEE Conference on Computer Vision and
  Pattern Recognition}, pages 175--185, 2018.

\bibitem{qiu2019deeplidar}
Jiaxiong Qiu, Zhaopeng Cui, Yinda Zhang, Xingdi Zhang, Shuaicheng Liu, Bing
  Zeng, and Marc Pollefeys.
\newblock Deeplidar: Deep surface normal guided depth prediction for outdoor
  scene from sparse lidar data and single color image.
\newblock In {\em Proceedings of the IEEE Conference on Computer Vision and
  Pattern Recognition}, pages 3313--3322, 2019.

\bibitem{ye2018depth}
Xinchen Ye, Xiangyue Duan, and Haojie Li.
\newblock Depth super-resolution with deep edge-inference network and
  edge-guided depth filling.
\newblock In {\em 2018 IEEE International Conference on Acoustics, Speech and
  Signal Processing (ICASSP)}, pages 1398--1402. IEEE, 2018.

\bibitem{scharstein2002taxonomy}
Daniel Scharstein and Richard Szeliski.
\newblock A taxonomy and evaluation of dense two-frame stereo correspondence
  algorithms.
\newblock {\em International journal of computer vision}, 47(1-3):7--42, 2002.

\bibitem{scharstein2014high}
Daniel Scharstein, Heiko Hirschm{\"u}ller, York Kitajima, Greg Krathwohl, Nera
  Ne{\v{s}}i{\'c}, Xi~Wang, and Porter Westling.
\newblock High-resolution stereo datasets with subpixel-accurate ground truth.
\newblock In {\em German conference on pattern recognition}, pages 31--42.
  Springer, 2014.

\bibitem{eldesokey2018propagating}
Abdelrahman Eldesokey, Michael Felsberg, and Fahad~Shahbaz Khan.
\newblock Propagating confidences through cnns for sparse data regression.
\newblock {\em arXiv preprint arXiv:1805.11913}, 2018.

\bibitem{liu2018image}
Guilin Liu, Fitsum~A Reda, Kevin~J Shih, Tingchun Wang, Andrew Tao, and Bryan
  Catanzaro.
\newblock Image inpainting for irregular holes using partial convolutions.
\newblock pages 89--105, 2018.

\bibitem{Gaidon:Virtual:CVPR2016}
A~Gaidon, Q~Wang, Y~Cabon, and E~Vig.
\newblock Virtual worlds as proxy for multi-object tracking analysis.
\newblock In {\em CVPR}, 2016.

\bibitem{eigen2014depth}
David Eigen, Christian Puhrsch, and Rob Fergus.
\newblock Depth map prediction from a single image using a multi-scale deep
  network.
\newblock In {\em Advances in neural information processing systems}, pages
  2366--2374, 2014.

\end{thebibliography}
\end{quote}
\end{document}